\documentclass[final]{cvpr}

\usepackage{times}
\usepackage{epsfig}

\usepackage{graphicx}
\usepackage{amsmath}
\usepackage{amssymb}
\usepackage{booktabs}
\usepackage[dvipsnames]{xcolor}

\newcommand{\mname}{UltraSR}

\usepackage{booktabs}
\usepackage{array}
\newcolumntype{L}[1]{>{\raggedright\let\newline\\\arraybackslash\hspace{0pt}}m{#1}}
\newcolumntype{C}[1]{>{\centering\let\newline\\\arraybackslash\hspace{0pt}}m{#1}}
\newcolumntype{R}[1]{>{\raggedleft\let\newline\\\arraybackslash\hspace{0pt}}m{#1}}
\usepackage{tablefootnote}
\usepackage{caption}
\usepackage{subcaption}
\usepackage{multirow}

\usepackage[pagebackref=true,breaklinks=true,colorlinks,bookmarks=false]{hyperref}

\begin{document}

\title{UltraSR: Spatial Encoding is a Missing Key for Implicit Image Function-based Arbitrary-Scale Super-Resolution}

\author{Xingqian Xu$^{1,3}$, Zhangyang Wang$^2$, Humphrey Shi$^{1,3}$\\
\\
{\small $^1$SHI Lab @ UIUC \& UO, $^2$UT Austin, $^3$Picsart AI Research (PAIR)}}

\maketitle

\begin{abstract}

The recent success of NeRF and other related implicit neural representation methods has opened a new path for continuous image representation, where pixel values no longer need to be looked up from stored discrete 2D arrays but can be inferred from neural network models on a continuous spatial domain. Although the recent work LIIF has demonstrated that such novel approaches can achieve good performance on the arbitrary-scale super-resolution task, their upscaled images frequently show structural distortion due to the inaccurate prediction of high-frequency textures. 
In this work, we propose \textbf{UltraSR}, a simple yet effective new network design based on implicit image functions in which we deeply integrated spatial coordinates and periodic encoding with the implicit neural representation. 
Through extensive experiments and ablation studies, we show that spatial encoding is a missing key toward the next-stage high-performing implicit image function. 
Our UltraSR sets new state-of-the-art performance on the DIV2K benchmark under all super-resolution scales compared to previous state-of-the-art methods. UltraSR also achieves superior performance on other standard benchmark datasets in which it outperforms prior works in almost all experiments.

\end{abstract}

\vspace{0.3cm}
\section{Introduction}
Image data has long been stored and computed with discrete 2D arrays, which is a compromised solution between flexibility and complexity, to some extent. The popularity of convolutional layers also strengthened discretization in which convolutional kernel parameters are scattered on fixed spatial locations. Despite some efforts to break such constraints (\eg RoI alignment~\cite{faster_rcnn}, deformable convolution~\cite{deformconv}, etc.), a majority of research works in computer vision accepted and adapted such discretization without question. Nevertheless, we must not underestimate the potential of utilizing the continuity of spatial domain in deep learning. The new state-of-the-art (SOTA) work LIIF~\cite{liif} on arbitrary-scale super-resolution (SR) suggests that we are capable of learning a continuous function representing high-resolution (HR) images using a multilayer perceptron (MLP) network with low-resolution (LR) features and pixel coordinates as its inputs. The highlight is that a simple MLP is good enough to generate high-quality SR images across all scales and beat prior SR networks trained on fixed scales. 

Despite the initial success of LIIF, we find plenty of structural distortions and other noticeable artifacts in SR images generated by LIIF. This observation motivates us to carefully examine the current ideas and methodology for arbitrary-scale SR approaches, which serves as our primary goal of this work. 

\begin{figure}[t!]
    \centering
    \includegraphics[width=0.48\textwidth]{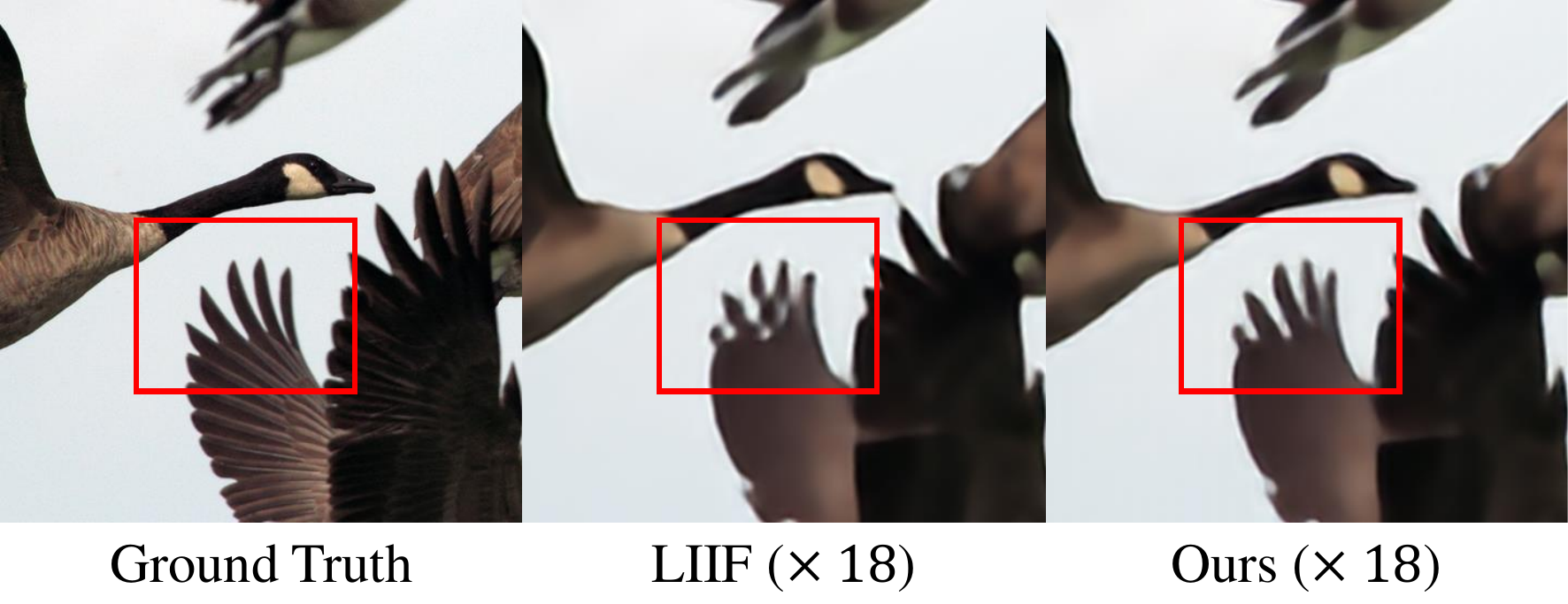}
    \caption{The quality comparison between LIIF~\cite{liif} and our \mname{} at resolution scale $\times 18$. As shown in the figure, our model can avoid structural distortions at extreme scales. }
\label{fig:highlight}
\end{figure}

Recent neural rendering works~\cite{nerf, derf, nerf++, nex} suggest that spatial encoding is critical for recovering high-frequency details in 3D scenes. 
Extending the idea into 2D tasks is natural, and a properly designed method should generate images carrying delicate details without losing spatial continuity.
In this paper, we will introduce a new arbitrary-scale SR method, namely \mname{}, which is a carefully designed network that learns the implicit image functions with all the merits mentioned above. 
With the help of our periodic spatial encoding and deep coordinate fusion, \mname{} can stably boost the performance in SR and surpasses the prior SOTA method LIIF on all resolution scales.

Meanwhile, arbitrary-scale SR is a rising research topic with tremendous application potentials. Prior CNN-based SR approaches usually apply to only one fixed upsampling scale and cannot discretionarily adjust their output dimension. 
Such design creates a huge gap between academic research and practical usage, resulting in most image applications relying heavily on bicubic interpolation despite its poor quality. 
Empowered by the rapidly advancing techniques in implicit neural representation, images and scenes can now be generalized by network-learned implicit functions on various vision topics (\eg free-viewpoint scene reconstruction, 3D video generation, etc.). Specifically for our SR task, the idea that uses a single network for all zoom-in scales will bring more convenience than ever to downstream users. 

In summary, the three main contributions of this article are as follows:

\begin{itemize}

    \item[$\bullet$] We reveal the importance of spatial encoding for implicit functions on 2D images through analysis and experiments. 
    
    \item[$\bullet$] We introduce a series of architecture designs such as deep coordinate fusion and residual MLP that work well with spatial encoding. Without these designs, the effectiveness of applying spatial encoding in implicit functions would be vastly reduced. 
    
    \item[$\bullet$] Our \mname{} sets the new SOTA on arbitrary-scale SR. The performance of \mname{} surpasses LIIF on the DIV2K validation set under all resolution scales. Meanwhile, \mname{} also beats prior arts on other five benchmark datasets under a majority of the testing schemes.
\end{itemize}

\vspace{0.1cm}
\section{Related Work}\label{sec:2}
This section will briefly introduce recent works on implicit neural representation, spatial encoding, and various super-resolution methods related to our work.

\vspace{0.1cm}
\subsection{Rendering with Spatial Encoding}

Learning 3D scene representation with a parameterized neural network has been largely explored by recent works from various angles such as implicit signed distance function~\cite{sdf1, sdf2, deepsdf}, occupancy~\cite{occ1, occ2, genova_shape2}, volume rendering (\ie radiance field)~\cite{nerf, nsvf, diffvol, derf, nerf++, nex}, and shapes~\cite{sal, genova_shape1, genova_shape2}. Such implicit neural representation also started to influence traditional 2D tasks such as image representation~\cite{hnet, siern}, super-resolution~\cite{liif}, and medical image analysis~\cite{nerd}. Among these works, spatial encoding has played a critical role in 3D scene reconstruction, whose effectiveness has not been fully explored in the 2D domain. In NeRF~\cite{nerf}, Mildenhall \etal expanded the input 3D coordinates with periodic functions before fed them into NeRF. They showed clear improvements on rendered images which were free from blurry details and structural distortion. Later 3D works~\cite{nsvf, nerf++, graf, nex}, whose aims were to reduce the rendering speed and improve output quality, involved such design without question. Other works~\cite{specbias, ssdgan, siern} analyzed spatial encoding on a theory level and brought out hypotheses for the tremendous performance boosts. In~\cite{specbias}, Rahaman \etal highlighted that regular neural nets had strong learning biases toward low-frequency spectral. Chen~\cite{ssdgan} also drawn a similar conclusion that generators tended to distort images in the high-frequency domain. Recently in SIERN~\cite{siern}, image gradients and laplacians were near-flawlessly reconstructed by replacing ReLU with periodic functions in neural networks.

\vspace{0.1cm}
\subsection{Single Image Super-Resolution}

Single image super-resolution (SISR) had decades of history in research. Traditional approaches can be roughly divided into patch-based~\cite{sisrt_pb1, sisrt_pb2}, edge-based~\cite{sisrt_eb1, sisrt_eb2}, and statistic-based~\cite{sisrt_sb1} methods. The first CNN-based SISR work was SRCNN~\cite{srcnn}, consisting of three convolutional layers representing patch extraction, feature mapping, and image reconstruction. Later, researchers proposed residual structures such as VDSR~\cite{vdsr}, IRCNN~\cite{ircnn}, and SRResNet~\cite{srresnet}. Lim \etal~\cite{edsr} proposed EDSR, in which they improved the residual blocks by removing BN layers. Yu \etal~\cite{wdsr} further enhanced EDSR into WDSR with an even wider channel before ReLU. Meanwhile, RDN~\cite{rdn} proposed residual dense block (RDB) with three densely connected convolutional layers that boosted SR quality to a new level. Recently, the non-local attention module~\cite{non_local, attn} becomes rather popular. Zhang \etal~\cite{rcan} adopted the non-local concept in RCAN, proposed a high-performing SR model with residual channel attention. Mei \etal~\cite{yiqun,mei2020pyramid} performed cross-scale attention to exhaustively search repeated patterns for SR.

\vspace{0.1cm}
\subsection{Arbitrary-Scale Super-Resolution}

\begin{figure*}[t!]
    \centering
    \includegraphics[width=0.9\textwidth]{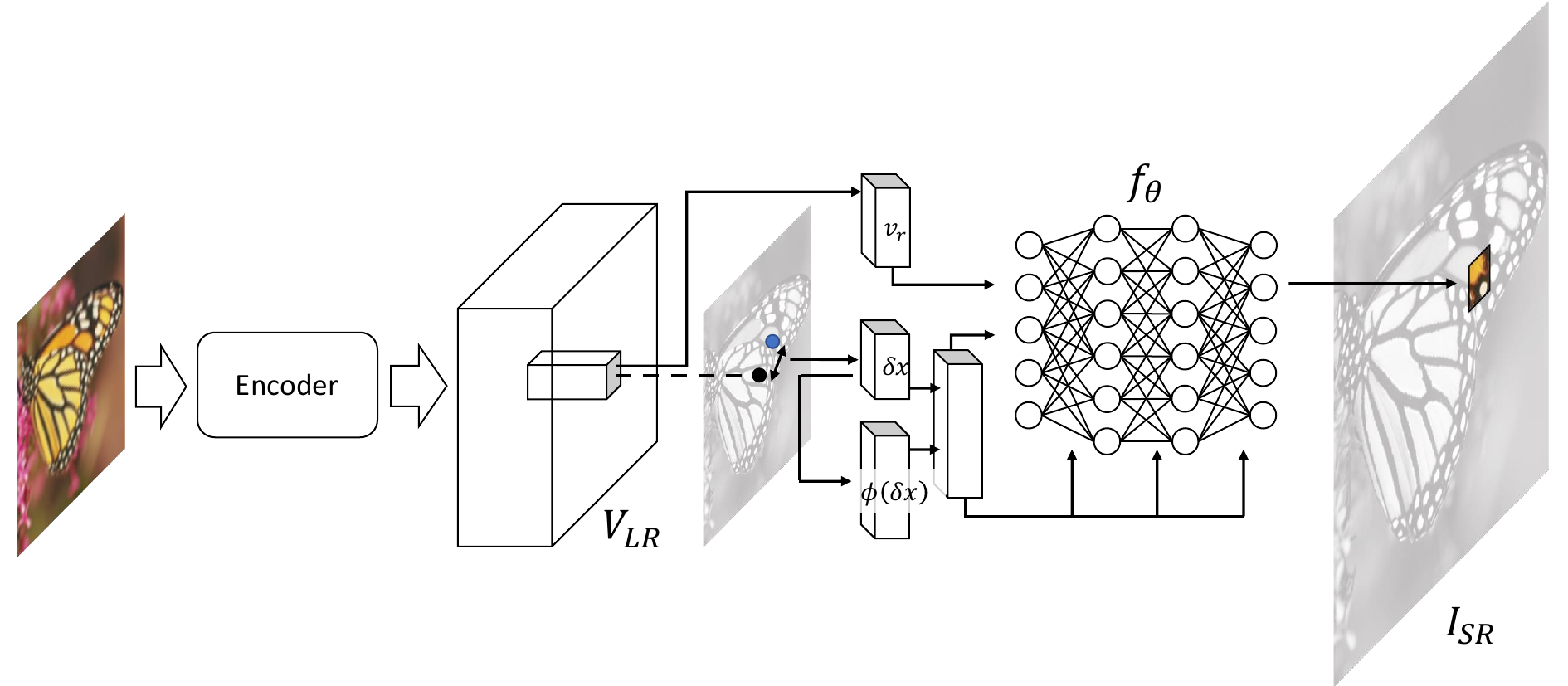}
    \caption{This figure shows the overall structure of our \mname{}. The blue point refers to the location of our target HR pixel that needs to be rendered. The black point shows the location of the nearby feature in the LR feature map $V_{LR}$. $\delta x$ is the location difference and $\phi(\cdot)$ is the spatial encoding function. $f_{\theta}$ is our rendering network in which coordinates and encoding are deeply fused to each network's hidden layer.}
\label{fig:network}
\end{figure*}

Arbitrary-scale SR methods significantly surpass previous SISR works in terms of practice and convenience. The idea that uses one neural network for any SR scale could be dated back to MDSR~\cite{edsr} introduced by Lim. MDSR proposed the pre-processing and scale-specific upsampling modules and placed them in the network's front and back to achieve SR in different scales. Nevertheless, MDSR could only handle scales other than $\times 2$, $\times 3$, and $\times 4$, and thus it was not an actual arbitrary-scale method. MetaSR~\cite{metasr} was the first CNN-based SR method on arbitrary scales. The proposed Meta-Upscale module mapped SR pixels onto the LR domain using the nearest-neighbor rule. During training, all mapped values multiplied dynamically learned weights based on scales and coordinates. The Meta-Upscale module then generated images with some additional convolutional layers. The recent SOTA work LIIF~\cite{liif} proposed a novel framework in which pixel values were computed by MLP using coordinates, cell size (\ie scale factors), and LR features. LIIF also showed robust performance under extreme SR scales up to $\times 30$. The new evaluation standard in LIIF that we test one network with an extensive range of scales will also help to bring more powerful SR models for practical usage.

\vspace{0.4cm}
\section{Methods}\label{sec:3}
In this session, we introduce \mname{}, a new arbitrary-scale SR model capable of generating any scales of SR images from LR images using implicit neural representation. Our work is strongly motivated by the recent works NeRF~\cite{nerf} and LIIF~\cite{liif}. The former demonstrated that combining neural rendering with spatial encoding can synthesize free-viewpoint 3D scenes with fine details, and the latter proved that one properly learned implicit image function could restore images with satisfactory quality at arbitrary SR scales. 

Like LIIF, we formulate the implicit form of any image in the HR domain with the following equation: 

\begin{equation}
    s = f_{\theta}(v_r, \delta x), \quad \delta x \propto x-x_r
    \label{eq:imp_func}
\end{equation}
 
\noindent where $s$ is the value of the target pixel in HR domain, $v_r$ is the feature vector of the reference location in LR domain, and $\delta x$ is the normalized distance between target pixel $x$ and the reference location $x_r$. The feature vector $v_r$ is extracted from the LR feature map $V_{LR} \in R^{C\times H \times W}$ in which its spacial location $x_r$ is near to $x$. $f_{\theta}$ is then the implicit image function simulated by a network with parameter $\theta$. 

\begin{figure}[h!]
\centering
\vspace{-0.1cm}
\begin{subfigure}{.23\textwidth}
  \centering
  \includegraphics[width=.98\linewidth]{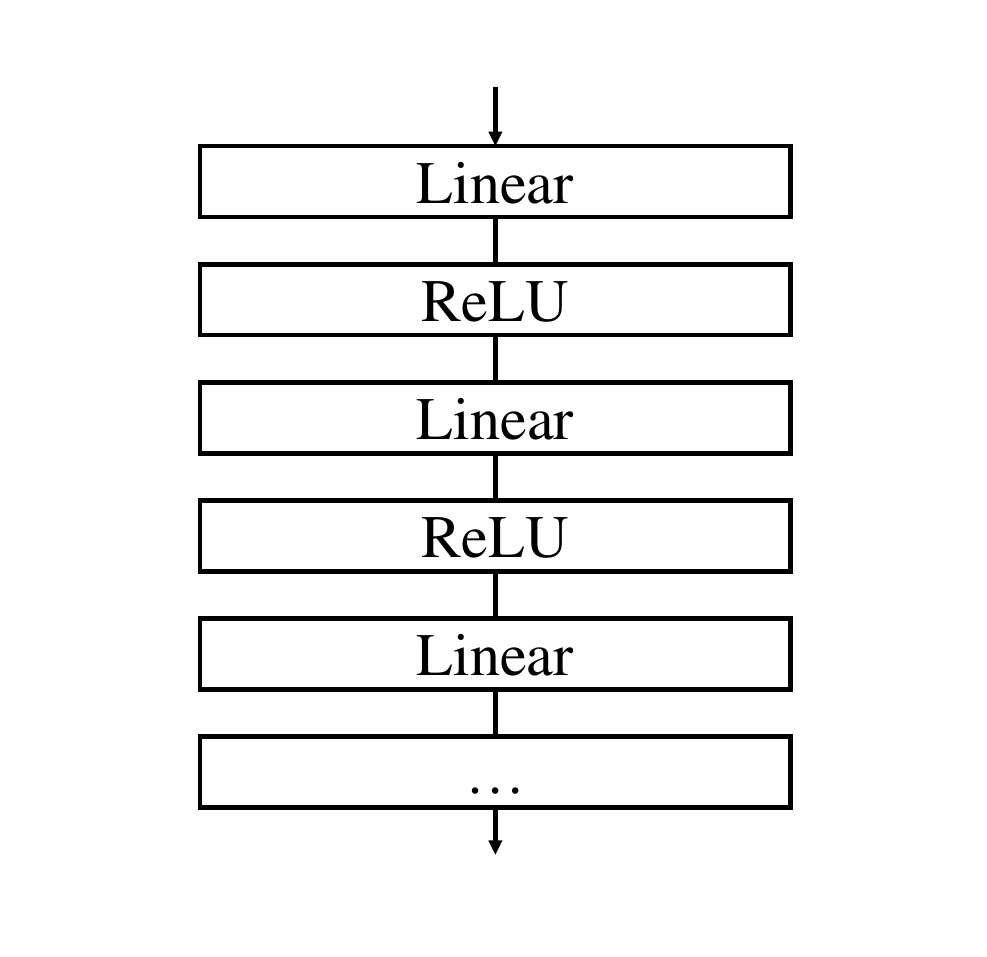} 
  \caption{MLP}
\end{subfigure}
\begin{subfigure}{.23\textwidth}
  \centering
  \includegraphics[width=.98\linewidth]{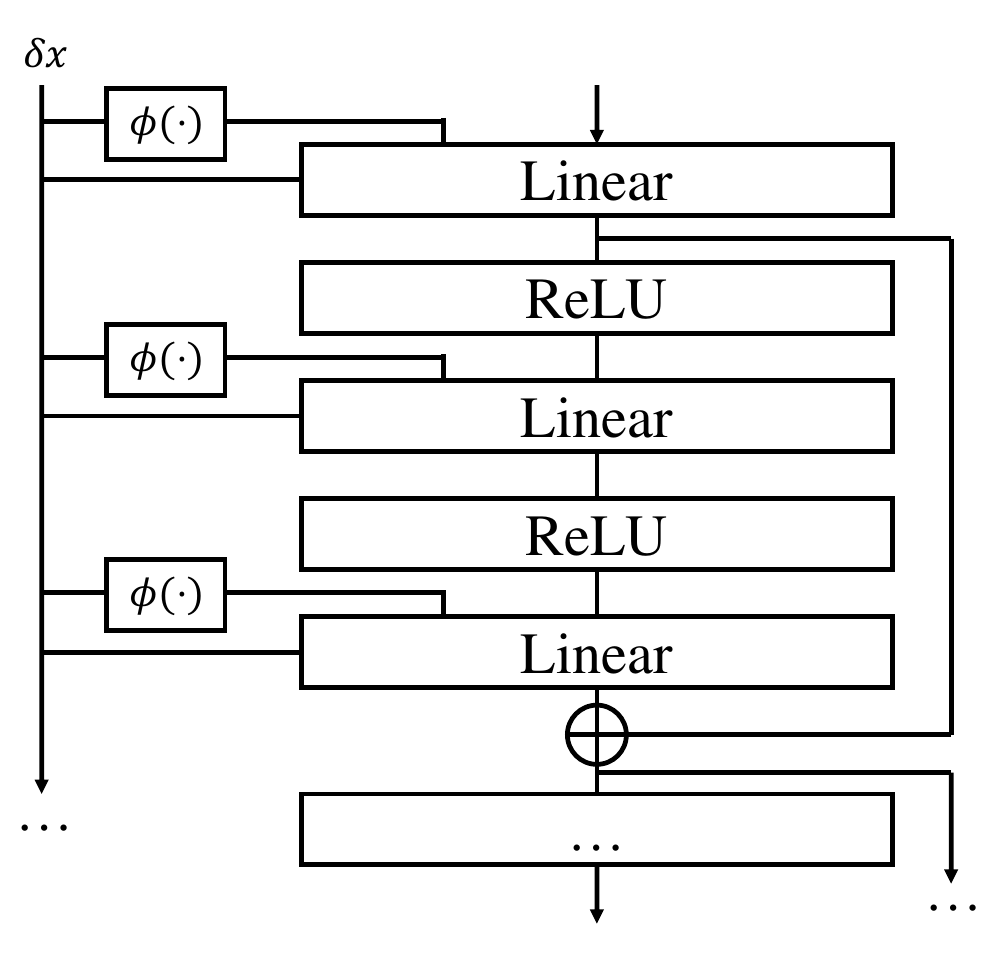}  
  \caption{ResMLP}
\end{subfigure}
\caption{This figure shows the structure of the vanilla MLP used by LIIF~\cite{liif} (left) and our ResMLP with coordinate fusion and residual links (right).}
\label{fig:resmlp}
\end{figure}

\subsection{Periodic Spatial Encoding}\label{sec:3_1}

\begin{figure*}[t!]
    \centering
    \includegraphics[width=0.99\textwidth]{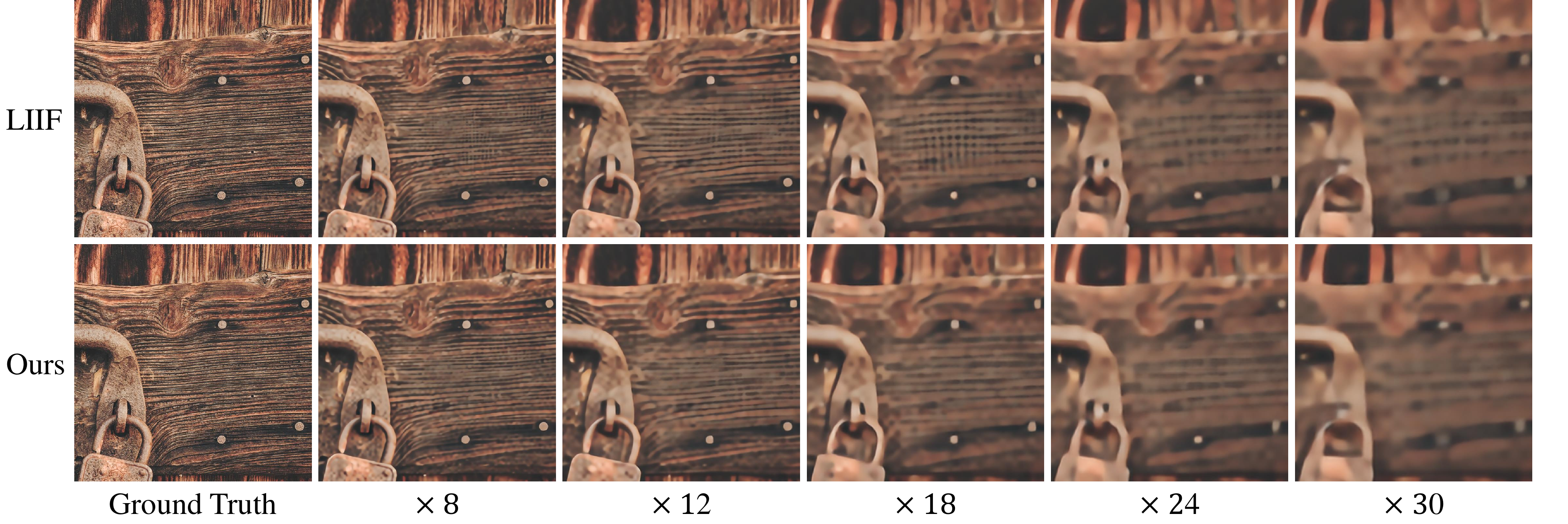}
    \caption{An example shows a group of persistent grid-like artifacts on all resolution scales generated by LIIF. Such artifacts do not appear in our \mname{}. Please zoom in for details.}
\label{fig:liif_vs_ours_detail}
\end{figure*}

Learning how to reconstruct the high-frequency part of the image is the key in the SR task. Many recent works~\cite{nerf, derf, nerf++, nex} have shown that a carefully designed spatial encoding can help the network recover fine details in 3D scenes. LIIF~\cite{liif} overlooked the importance of these spatial encoding by directly feeding coordinates into the implicit image function represented by a vanilla MLP. We empirically notice that, without spatial encoding, these neural representations tend to generate images with structural distortions and other noticeable artifacts. Moreover, these distortions and artifacts are persistent across different SR scales (see Figure~\ref{fig:liif_vs_ours_detail}), which hinder real-world applications. We believe that the phenomenon aligns with the recent discovery in~\cite{specbias, ssdgan} that neural networks are biased towards low-frequency signals and are insensitive to high-frequency signals. Therefore, in order to enhance the network in the high-frequency domain, our \mname{} expands the regular 2D linear spatial input with 48D periodic spatial encoding under the following equations:

\begin{equation}\begin{aligned}
    \phi(x) &= (sin(w_1x), cos(w_1x), \\
    &\quad\quad\quad\quad sin(w_2x), cos(w_2x), ... )
\label{eq:positional_encoding}
\end{aligned}\end{equation}

\begin{equation}
    s = f_{\theta}\left(v_r, \delta x, \phi(\delta x)\right)\\
\label{eq:imp_func2}
\end{equation}

\noindent where the frequency parameters $w_1, w_2, ...$ are initially set to $2e^{n}, n\in 1, 2, ...$, and are later fine-tuned in training. We follow the convention and use $cos$ and $sin$ as our encoding basis. At last, the input coordinates $\delta x$ together with their spatial encoding $\phi(\delta x)$ are feed into the implicit image function in \mname{} as shown in Equation~\ref{eq:imp_func}. 

\vspace{0.1cm}
\subsection{Deep Coordinate Fusion}

Nevertheless, a simple concatenation of coordinates and encoding does not provide us a solution. We find no obvious evidence that the quality of the output images can be immediately improved if we feed these spatial encodings into the vanilla MLP used in LIIF~\cite{liif}. It is also skeptical because recent neural rendering articles heavily rely on MLP structures, but few have investigated an optimal network structure. We have found that MLP is somewhat sub-optimal in our SR task because it does not prioritize the network inputs based on their importance. 

As shown in Equation~\ref{eq:imp_func2}, the three inputs of our implicit image representation are feature vectors, coordinates, and spatial encoding. We believe that the 2D coordinates and their encoding are more important than the feature vectors. Such an assumption is based on the fact that the feature vector in LR cannot differentiate HR pixels within the nearby $k^2$ region with an upsampling scale $k$. Therefore, all detail textures in the SR images must highly depend on the pixel coordinates and the spatial encoding. Therefore, we adjust our network so that these inputs are integrated with our MLP much tighter. Figure~\ref{fig:network} shows the overall structure of \mname{}, in which we concatenate the 2D coordinates with the 48D encoding and feed them to all hidden layers. Such fusion ensures that all hidden layers can directly access the output pixels' critical spatial information and utilize the high-frequency hints inside the spatial encoding off-the-shelf.

\begin{table*}[t!]
\centering
\resizebox{0.95\textwidth}{!}{
    \begin{tabular}{
            R{3.5cm}
            |C{1.1cm}
            C{1.1cm}
            C{1.1cm}
            |C{1.1cm}
            C{1.1cm}
            C{1.1cm}
            C{1.1cm}
            C{1.1cm}}
        \toprule
            Method 
            & $\times 2$
            & $\times 3$
            & $\times 4$
            & $\times 6$
            & $\times 12$
            & $\times 18$
            & $\times 24$
            & $\times 30$
            \\
        \midrule
        Bicubic
            & 31.01
            & 28.22
            & 26.66
            & 24.82
            & 22.27
            & 21.00
            & 20.19
            & 19.59
            \\
        EDSR-baseline
            & 34.55
            & 30.90
            & 28.92
            & --
            & --
            & --
            & --
            & --
            \\
        MetaSR-EDSR
            & 34.64
            & 30.93
            & 28.92
            & 26.61
            & 23.55
            & 22.03
            & 21.06
            & 20.37
            \\
        LIIF-EDSR
            & 34.67
            & 30.96
            & 29.00
            & 26.75
            & 23.71
            & 22.17
            & 21.18
            & 20.48
            \\
        \mname{}-EDSR (ours)
            & \textbf{34.69}
            & \textbf{31.02}
            & \textbf{29.05}
            & \textbf{26.81}
            & \textbf{23.75}
            & \textbf{22.21}
            & \textbf{21.21}
            & \textbf{20.51}
            \\
        \midrule
        MetaSR-RDN
            & \textbf{35.00}
            & 31.27
            & 29.25
            & 26.88
            & 23.73
            & 22.18
            & 21.17
            & 20.47
            \\
        LIIF-RDN
            & 34.99
            & 31.26
            & 29.27
            & 26.99
            & 23.89
            & 22.34
            & 21.31
            & 20.59
            \\
        \mname{}-RDN (ours)
            & \textbf{35.00}
            & \textbf{31.30}
            & \textbf{29.32}
            & \textbf{27.03}
            & \textbf{23.93}
            & \textbf{22.36}
            & \textbf{21.33}
            & \textbf{20.61}
            \\
        \bottomrule
    \end{tabular}
}
\vspace{0.2cm}
\caption{
    PSNR (dB) comparison between MetaSR~\cite{metasr}, LIIF~\cite{liif}, and \mname{} (ours) on the DIV2K validation set with different SR scales. All three methods use one model for all scales. The bold numbers indicate the best results. 
}
\vspace{0.3cm}
\label{table:perf_div2k}
\end{table*}

\begin{table*}[t!]
\centering
\resizebox{0.92\textwidth}{!}{
    \begin{tabular}{
            C{2.5cm}
            |C{3.5cm}
            |C{1.1cm}
            C{1.1cm}
            C{1.1cm}
            |C{1.1cm}
            C{1.1cm}
            C{1.1cm}}
        \toprule
            Dataset
            & Method 
            & $\times 2$
            & $\times 3$
            & $\times 4$
            & $\times 6$
            & $\times 8$
            & $\times 12$
            \\
        \midrule
        \multirow{4}{*}{Set5}
        & RDN
            & \textbf{38.24}
            & \textbf{34.71}
            & 32.47
            & --
            & --
            & --
            \\
        & MetaSR-RDN
            & 38.22
            & 34.63
            & 32.38
            & 29.04
            & 29.96
            & --
            \\
        & LIIF-RDN
            & 38.17
            & 34.
            & \textbf{32.50}
            & 29.15
            & 27.14
            & \textbf{24.86}
            \\
        & \mname{}-RDN (ours)
            & 38.21
            & 34.67
            & 32.49
            & \textbf{29.33}
            & \textbf{27.24}
            & 24.81
            \\
        \midrule
        \multirow{4}{*}{Set14}
        & RDN
            & \textbf{34.01}
            & 30.57
            & 28.81
            & --
            & --
            & --
            \\
        & MetaSR-RDN
            & 33.98
            & 30.54
            & 28.78
            & 26.51
            & 24.97
            & --
            \\
        & LIIF-RDN
            & 33.97
            & 30.53
            & 28.80
            & 26.64
            & 25.15
            & 23.24
            \\
        & \mname{}-RDN (ours)
            & 33.97
            & \textbf{30.59}
            & \textbf{28.86}
            & \textbf{26.69}
            & \textbf{25.25}
            & \textbf{23.32}
            \\
        \midrule
        \multirow{4}{*}{B100}
        & RDN~
            & 32.34
            & 29.26
            & 27.72
            & --
            & --
            & --
            \\
        & MetaSR-RDN
            & 32.33
            & 29.26
            & 27.71
            & 25.90
            & 24.83
            & --
            \\
        & LIIF-RDN
            & 32.32
            & 29.26
            & 27.74
            & 25.98
            & 24.91
            & 23.57
            \\
        & \mname{}-RDN (ours)
            & \textbf{32.35}
            & \textbf{29.29}
            & \textbf{27.77}
            & \textbf{26.01}
            & \textbf{24.96}
            & \textbf{23.59}
            \\
        \midrule
        \multirow{4}{*}{Urban100}
        & RDN
            & 32.89
            & 28.80
            & 26.61
            & --
            & --
            & --
            \\
        & MetaSR-RDN
            & 32.92
            & 28.82
            & 26.55
            & 23.99
            & 22.59
            & --
            \\
        & LIIF-RDN
            & 32.87
            & 28.82
            & 26.68
            & 24.20
            & 22.79
            & 21.15
            \\
        & \mname{}-RDN (ours)
            & \textbf{32.97}
            & \textbf{28.92}
            & \textbf{26.78}
            & \textbf{24.30}
            & \textbf{22.87}
            & \textbf{21.20}
            \\
        \midrule
        \multirow{4}{*}{Manga109}
        & RDN
            & 39.18
            & 34.13
            & 31.00
            & --
            & --
            & --
            \\
        & MetaSR-RDN
            & --
            & --
            & --
            & --
            & --
            & --
            \\
        & LIIF-RDN
            & \textbf{39.26}
            & 34.21
            & 31.20
            & 27.33
            & 25.04
            & 22.36
            \\
        & \mname{}-RDN (ours)
            & 39.09
            & \textbf{34.28}
            & \textbf{31.32}
            & \textbf{27.42}
            & \textbf{25.12}
            & \textbf{22.42}
            \\
        \bottomrule
    \end{tabular}
}
\vspace{0.2cm}
\caption{
    PSNR (dB) comparison between RDN~\cite{rdn}, LIIF~\cite{liif}, and \mname{} (ours) on five benchmark datasets. The performance of \mname{} surpasses RDN and LIIF in the majority of the table entries. Specifically, when the dataset is large (\eg B100, Urban100, Manga109), our results surpass prior works on all scales that are larger than 2.
}
\label{table:perf_benchmarkds}
\end{table*}

\vspace{0.1cm}
\subsection{Network details}

Besides the spatial encoding and the coordinate fusion, we add residual links in MLP, forming up residual-MLP (ResMLP) to strengthen its ability to generate images with fine SR details. The structure of our ResMLP follows the convention in ~\cite{resnet, srresnet, biggan}, in which hidden features are skip-connected between every two layers before activation (see Figure~\ref{fig:resmlp}). We have noticed that the residual link can help the network restore images with high fidelity. It frees the network from processing low-frequency information by passing them directly to the latter layers. Besides, we also adopt the same feature unfolds, local ensembling, and cell decoding mentioned in ~\cite{liif}. We will show results and comparisons on \mname{} and prior works in the next session.

\vspace{0.2cm}
\section{Experiment}\label{sec:4}
In this section, we will reveal the dataset, metric, and training details of our experiments. We will then discuss our model performance by comparing \mname{} with prior works. Lastly, we will further analyze \mname{} via several ablation studies.

\subsection{Dataset and Metrics}\label{sec:4_1}

The main dataset we use to train and evaluate our \mname{} is the DIV2K dataset~\cite{div2k} from NTIRE 2017 Challenge. DIV2K consists of 1000 2K high-resolution images together with the bicubic down-sampled low-resolution images under scale $\times 2$, $\times 3$ and $\times 4$. We maintain its original train validation split, in which we use the 800 images from the train set in training and the 100 images from the validation set for testing. Follows many prior works~\cite{edsr, srresnet, yiqun, rdn, metasr, liif}, we also report our model performance on 5 benchmark datasets: Set5~\cite{set5}, Set14~\cite{set14}, B100~\cite{b100}, Urban100~\cite{urban100} and Manga109~\cite{manga109}.

Meanwhile, we use the widely adopted Peak Signal-to-Noise Ratio (PSNR) as our evaluation metric on \mname{}. With pixel values ranging from 0 to 1, PSNR is computed as 10 multiplies the log10 of one over the mean square error between two images. Please also see Equation~\ref{eq:psnr} for PSNR computation in detail.

\begin{equation}
    PSNR = 10 \log_{10}\left(
        \frac{1}{\frac{1}{H\times W} \left\lVert I_{SR} - I_{HR} \right\rVert_2^2}
    \right)
\label{eq:psnr}
\end{equation}

\begin{figure*}[h!]
    \centering
    \includegraphics[width=0.85\textwidth]{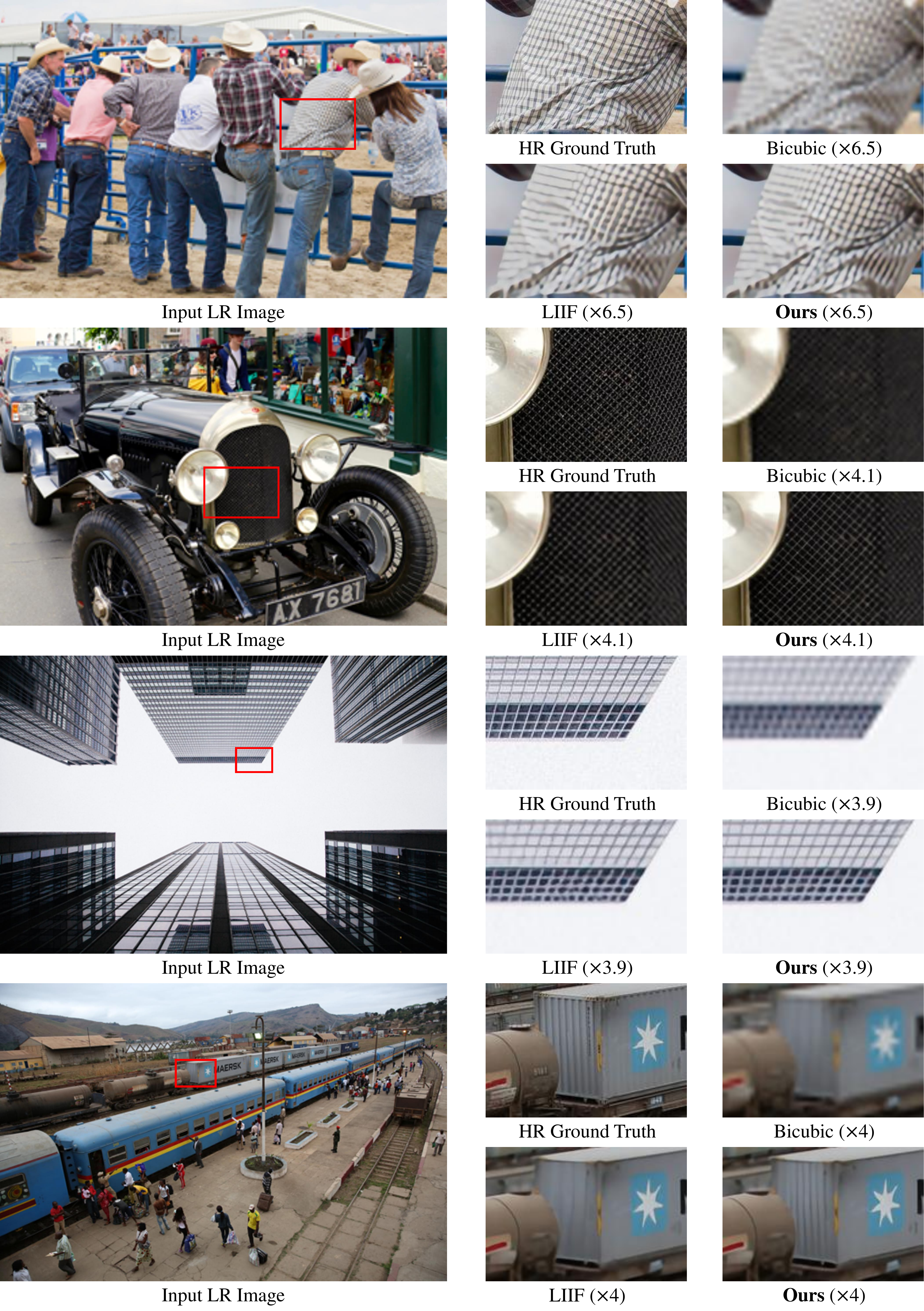}
    \caption{Additional SR images generated by LIIF~\cite{liif} and by our \mname{}. As shown in this figure, our model can avoid structural distortions, reduce artifacts and generate sharp edges in various types of scenes. Please zoom in for details.}
\label{fig:liif_vs_ours_large}
\end{figure*}

\vspace{0.2cm}
\subsection{Training Details}

To train \mname{}, we first create LR training images from ground truth HR images through bicubic interpolation. The downsampling scales of these LR images are uniformly sampled from $\times 2$ to $\times 4$. We then randomly crop 48x48 patches from these LR images and feed them into \mname{}'s encoder. After that, we randomly render 2304 pixels in the HR domain and back-propagate its l1 loss against the ground truth. For the optimizer, we choose ADAM with betas 0.9 and 0.999. We train the entire pipeline for 1000 epochs, and each epoch has 1000 iterations. Our initial learning rate is $10^{-4}$, and it decays by one-half at epoch 200, 400, 600, and 800. Except for some minor changes, the entire setting closely follows the convention in~\cite{liif, metasr}. Like LIIF~\cite{liif}, we also choose EDSR~\cite{edsr} and RDN~\cite{rdn} excluding upsampling layers as two choices of our encoder in \mname{}. EDSR is a compact-sized model, so we train our \mname{}-EDSR on one RTX 2080 Ti GPU with batch-size 16. Meanwhile, RDN contains more layers, so we train \mname{}-RDN on two RTX 2080 Ti GPUs with batch-size 8 per GPU.

\vspace{0.2cm}
\subsection{Results and Comparison}

Table~\ref{table:perf_div2k} compares the performances of MetaSR~\cite{metasr}, LIIF~\cite{liif} with our \mname{} under different upsampling scales on DIV2K. We notice that our model's PSNR results are consistently higher than MetaSR and LIIF using both encoders and on all scales. We also notice that these increments are maximized around scale $\times 3$ to $\times 12$, reach a bold 0.05 in PSNR. On the other hand, the PSNR increments are not so obvious at extreme scales when we provide either too much or too little LR information to the model.

In Table~\ref{table:perf_benchmarkds}, we also compare \mname{} with prior works on five standard benchmark datasets: Set5~\cite{set5}, Set14~\cite{set14}, B100~\cite{b100}, Urban100~\cite{urban100} and Manga109~\cite{manga109}. Similar to DIV2K, we demonstrate that \mname{} is a better solution by surpassing both RDN and LIIF in most of our experiment schemes. Specifically, on large datasets (\eg B100, Urban100, and Manga109) with scales larger than 2, our model surpasses RDN and LIIF in all experiments, even that RDN was trained one dedicated model for each scale. 

Lastly, we show the qualitative comparisons between LIIF and \mname{}. As shown in Figure~\ref{fig:liif_vs_ours_detail}, despite it can interpolate images on extreme scales, LIIF is more likely to distort high-frequency patterns and to generate images with low fidelity. with the help of both spatial encoding and coordinate fusion, the images generated by \mname{} are protected from these unwanted distortions. More examples can be found in Figure~\ref{fig:liif_vs_ours_large}.

\vspace{0.2cm}
\subsection{Ablation Studies}

In this session, we performed a series of experiments to justify the effectiveness of the spatial encoding and our network designs.

\begin{figure}[h!]
    \centering
    \includegraphics[width=0.5\textwidth]{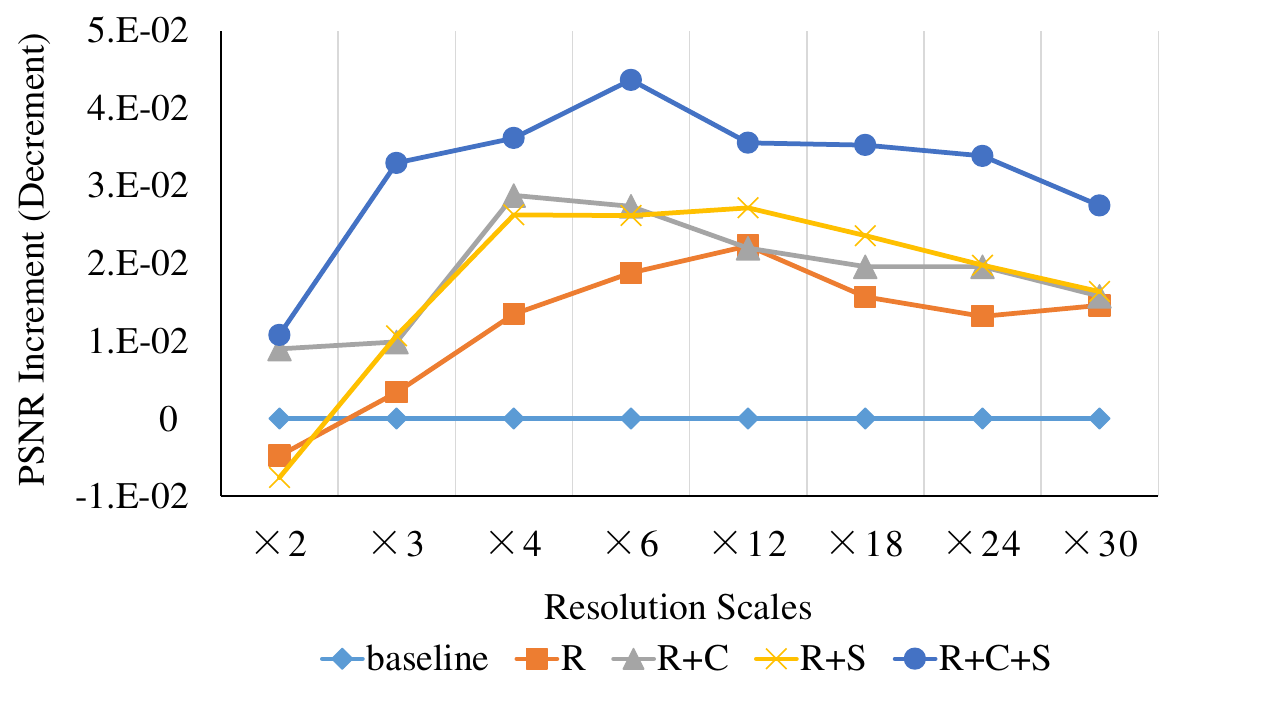}
    \caption{This plot shows PSNR increment/decrement on our model's performance against baseline under different settings. The baseline is plain \mname{} without spatial encoding(S), coordinate fusion(C), or residual links(R). R+C+S is the full \mname{}-EDSR shown in Table~\ref{table:perf_div2k}, and the rest are with parts of our design added to the baseline.}
\label{fig:aba}
\end{figure}

In our first ablation study, we trained multiple models, among which we progressively added the spacial encoding(S), coordinate fusion(C), and residual links(R). We used EDSR as our encoder in all models for a fair comparison. All other training parameters were kept unchanged. The results shown in Table~\ref{fig:aba} demonstrate that the simple spatial encoding does not work very well, but spatial encoding with coordinate fusion provides the best solution. We also show that ResMLP is a simple but better design than vanilla MLP in terms of accuracy.

\begin{figure}[t!]
    \centering
    \includegraphics[width=0.5\textwidth]{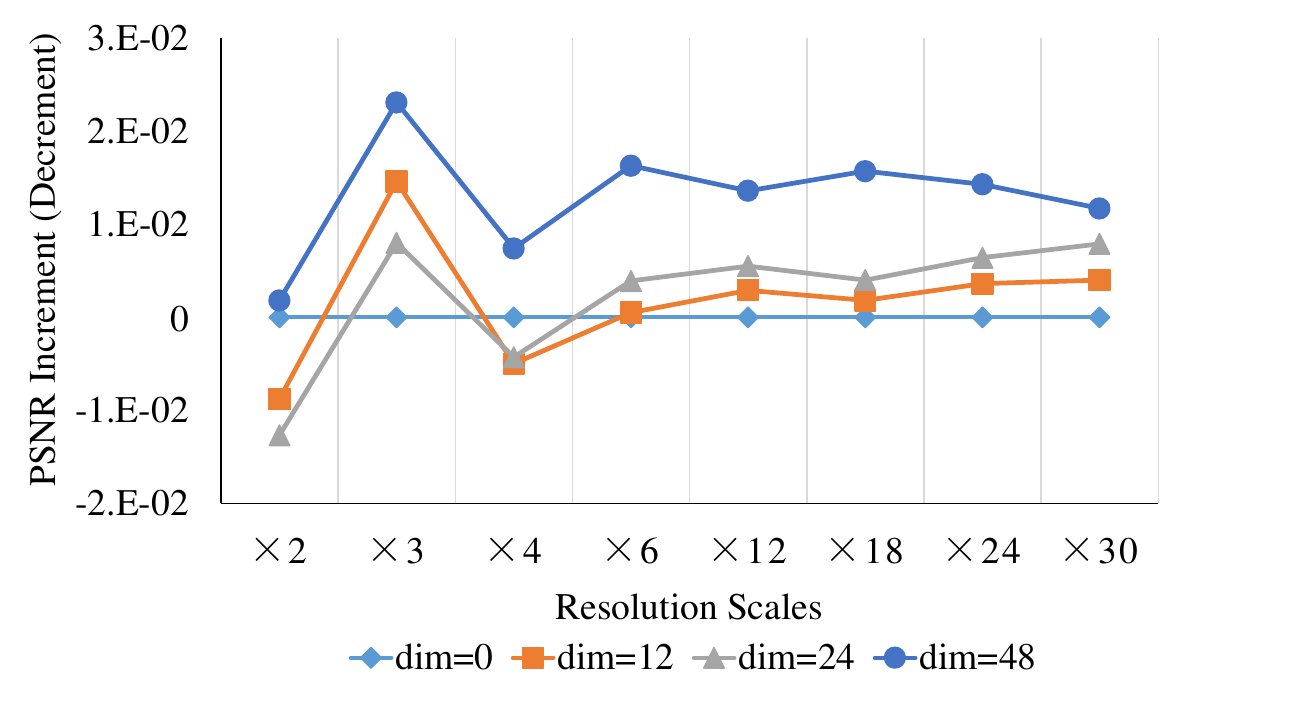}
    \caption{This plot shows the performance changes using different dimensions in our spatial encoding. The baseline is the model with dimension equals 0 (\ie the R+C model in Figure~\ref{fig:aba}). As shown, though there are some fluctuations in small scales, the growing encoding dimension can robustly increase the performance on large scales. }
\label{fig:abb}
\end{figure}

Our second study investigated how the dimension number of the spatial encoding influenced the model's performance. We tested the model with a total of three variations in the encoding dimension: 12, 24, 48. Other hyper-parameters were kept unchanged. The results in Figure~\ref{fig:abb} showed that: besides scale 2 and scale 4, the performance stably increased along with the dimension. The irregular behavior at scale 2 and scale 4 might be due to the 2 to 4 uniform scale sampling in the training phase data augmentation, which yield to a best result at scale 3.

\begin{figure}[h!]
    \centering
    \includegraphics[width=0.5\textwidth]{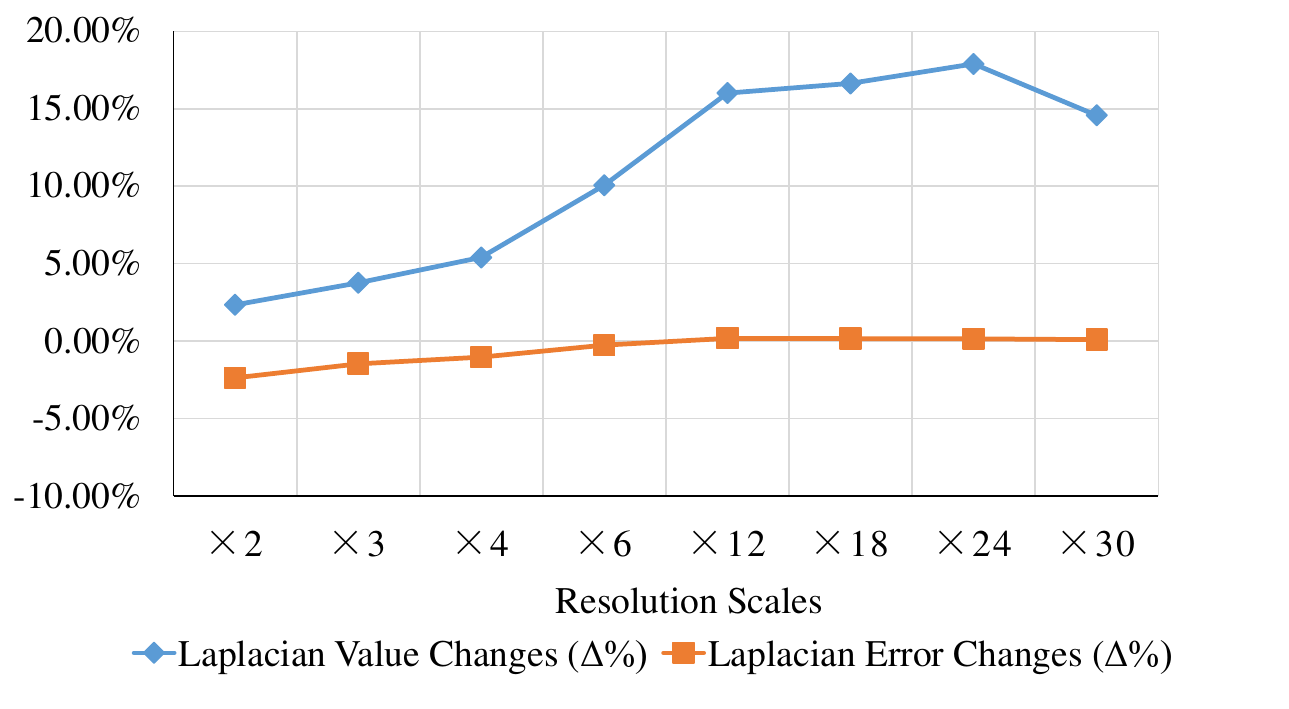}
    \caption{This plot shows the delta percentage changes of the mean absolute value/error of image Laplacian with/without our spatial encoding. As shown, more sharp edges are generated when using spatial encoding because the mean absolute value increase by averagely 10\% (blue line), while fewer errors are made because the mean absolute error against ground truth decreased by averagely 1\% (orange lines). In conclusion, spatial encoding can help to create more clear and reliable SR images.}
\label{fig:abc}
\end{figure}

Lastly, we proved that periodic spatial encoding could vastly improve the sharpness of the generated images without losing fidelity. We analyzed the Laplacian of the images generated by two models, one without spatial encoding (\mname{}-S) and the other with spatial encoding (\mname{}+S). These models were the same R+C and R+C+S models in Figure~\ref{fig:aba}. Laplacian filter is a well-known edge detector whose output represents the likelihood of an image patch contains an edge. Our study compared the delta percentage of \mname{}+S over \mname{}-S on two values across different scales: a) mean absolute value on Laplacian, and b) mean absolute error on Laplacian against ground truth. The former tells how willing the model generated sharp edges, and the latter tells how right these edges were. In Figure~\ref{fig:abc}, we showed that when comparing \mname{}+S with \mname{}-S, its mean absolute value on Laplacian is roughly 10\% higher, but the error is roughly 1\% lower. It proved that \mname{} with spatial encoding was capable of generating sharper images with fewer errors.

\vspace{0.3cm}
\section{Discussion}
In this section, we will discuss the limitations of \mname{} along with three potential future directions that can be extended from our work. 

\vspace{0.2cm}
\textbf{Encoding function space.} The Fourier basis (\ie $sin$ and $cos$) may be one type of many basis functions suitable for encoding spatial information. Our community lacks research on other types of basis in function space that may work better than the broadly applied Fourier basis. Examples such as radial and wavelet basis functions can be strong candidates to encode spatial information. We can as well promote task-specific basis functions based on different application scenarios.

\vspace{0.2cm}
\textbf{Tasks beyond SR.} We can extend the concept of implicit image functions to other 2D vision tasks beyond SR. For example, we may extend our framework to discriminative tasks such as classification~\cite{resnet}, detection~\cite{faster_rcnn}, and segmentation~\cite{deeplabv3p, daffnet, textseg, alignseg}. By far, we do not know much about such extension, but the generality of implicit functions suggests that we should have much broader research directions in computer vision.

\vspace{0.2cm}
\textbf{Perceptual-orientated SR.} Since SR becomes a rather ill-posed problem on extreme resolution scales, one should think about whether we can train a network to simulate image functions in a perceptual-orientated way, from which we can create photo-realistic images under extreme scales. The idea that combines \mname{} or any implicit neural representation works with GAN~\cite{gan} is attractive. It may become the visible next-stage research for neural rendering as well.

\vspace{0.3cm}
\section{Conclusions}
We introduce a novel arbitrary-scale super-resolution model \mname{} that deeply combines spatial encoding with implicit image function. We also reveal the importance of spatial encoding and coordinate fusion through result comparisons and visual evidence in which structural distortions can be effectively reduced. In conclusion, the PSNR performance of our \mname{} surpasses all prior arts on DIV2K dataset under all resolution scales. We also show our results on other 5 benchmark datasets from which the superiority of using spatial encoding can be once again demonstrated.

\clearpage

{\small
\bibliographystyle{ieee_fullname}
\bibliography{egbib}
}


\end{document}